\pdfoutput=1

\documentclass[11pt]{article}

\usepackage[preprint]{acl}
\usepackage{booktabs}
\usepackage{multirow}
\usepackage{times}

\usepackage{latexsym}
\usepackage{amsmath}
\usepackage[T1]{fontenc}
\usepackage[utf8]{inputenc}

\usepackage{microtype}

\usepackage{inconsolata}
\usepackage{wrapfig}
\usepackage{graphicx}

\title{Investigating the Impact of LLM Personality on Cognitive Bias Manifestation in Automated Decision-Making Tasks}

\author{Jiangen He \\
  School of Information Sciences \\The University of Tennessee, Knoxville \\
  \texttt{jiangen@utk.edu} \\\And
  Jiqun Liu \\
  School of Library and Information Studies \\ University of Oklahoma \\
  \texttt{jiqunliu@ou.edu} \\}

\newcommand{\EffectColor}[1]{
    \ifdim #1 pt > 0pt
        \cellcolor{green!#10} #1
    \else
        \cellcolor{red!#1-10} #1
    \fi
}
\begin{document}
\maketitle
\begin{abstract}
Large Language Models (LLMs) are increasingly used in decision-making, yet their susceptibility to cognitive biases remains a pressing challenge. This study explores how personality traits influence these biases and evaluates the effectiveness of mitigation strategies across various model architectures. Our findings identify six prevalent cognitive biases, while the sunk cost and group attribution biases exhibit minimal impact. Personality traits play a crucial role in either amplifying or reducing biases, significantly affecting how LLMs respond to debiasing techniques. Notably, Conscientiousness and Agreeableness may generally enhance the efficacy of bias mitigation strategies, suggesting that LLMs exhibiting these traits are more receptive to corrective measures. These findings address the importance of personality-driven bias dynamics and highlight the need for targeted mitigation approaches to improve fairness and reliability in AI-assisted decision-making.
\end{abstract}

\section{Introduction}
\begin{figure}[t]
    \centering
    \includegraphics[width=\linewidth]{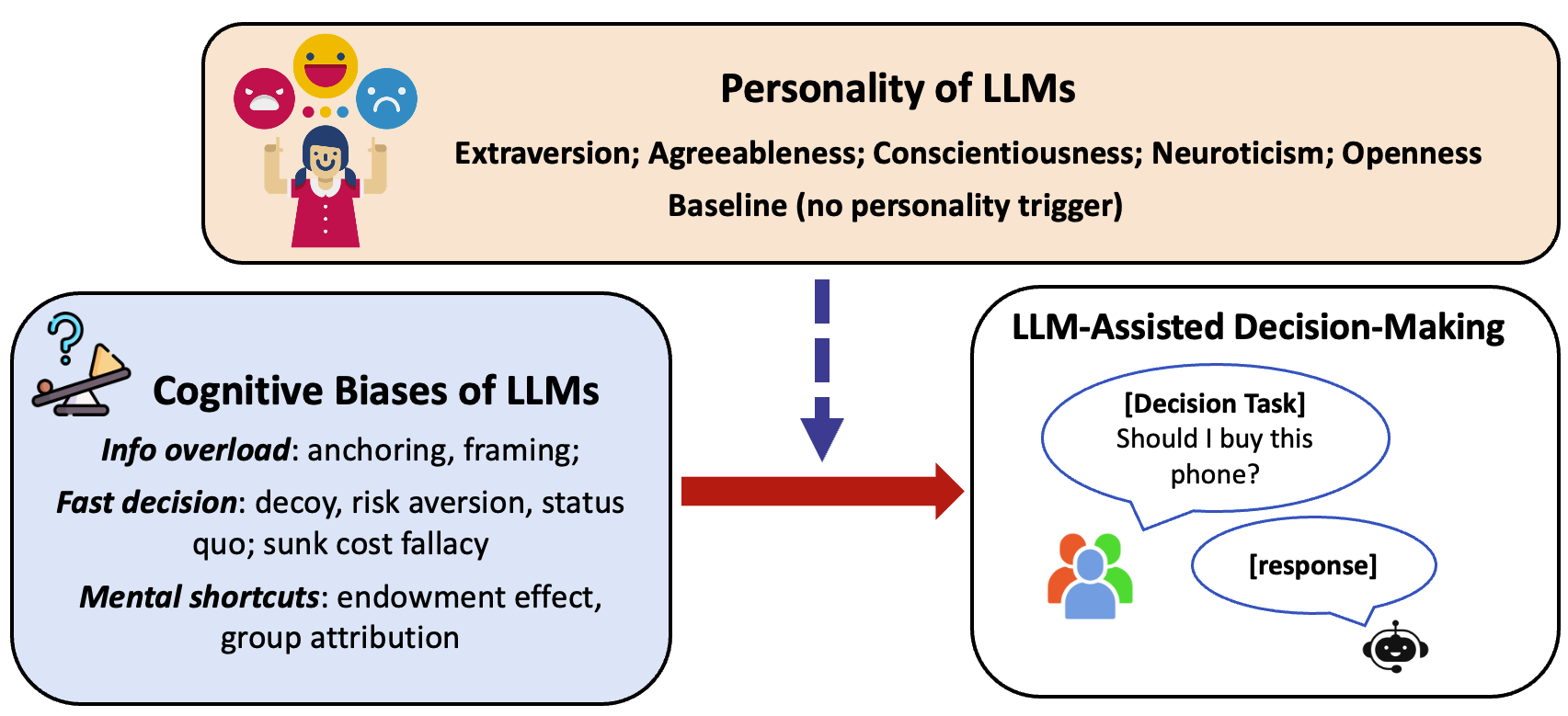}
    \caption{Personality-Bias Framework.}
    \label{fig:illustrate1}
\end{figure}

The rise of large language models (LLMs) has transformed decision-making processes across diverse domains, from education and finance to healthcare and policy. As these models increasingly take on roles traditionally held by human experts, concerns about their \textit{susceptibility to cognitive biases} have grown~\cite{hager2024evaluation, li2022pre}. While prior research has explored biases in AI-driven decision-making, a critical yet understudied factor is the role of \textit{LLM personality} in shaping these biases. Emerging evidence suggests that LLMs, much like humans, can exhibit distinct personality traits that influence how they process information, assess uncertainty, and generate recommendations~\cite{chen2024ai, liu2024decoy}. This raises an urgent question: \textit{Do LLM personalities amplify or mitigate cognitive biases in decision-making?} Addressing this question is essential for ensuring that AI-assisted tasks remains reliable and free from unintended distortions. This open challenge, illustrated in Figure~\ref{fig:illustrate1}, motivates our evaluation study on LLM reported here. 

\subsection{Cognitive Biases in Decision-Making}
\textit{Cognitive biases} are systematic deviations from rational judgment that significantly influence human judgments and decision-making outcomes~\cite{kahneman2003maps, liu2023behavioral}. Extensive research in Psychology and Behavioral Economics has identified numerous such biases across varying decision settings, such as anchoring bias, confirmation bias, decoy effect, and framing effect, which affect how individuals process information, perceive available options, assess utility and make choices~\cite{benartzi2007heuristics, tversky1992advances}. For instance, the anchoring bias leads people to rely heavily on the first piece of information encountered when making decisions~\cite{furnham2011literature}, while the decoy effect occurs when the presence of an asymmetrically dominated option influences a person's preference between two other choices, often leading to irrational decisions~\cite{chen2024decoy, wedell1996using}. These biases can result in suboptimal decisions and biased judgments across critical contexts, from financial investments to healthcare choices. Beyond traditional decision-making scenarios, researchers also found that users' cognitive biases affect their interactions with interactive information systems of varying modalities and shape their judgments on retrieved and generated information~\cite[e.g.][]{liu2023behavioral, ji2024towards, azzopardi2021cognitive, lin2023mind, chen2023reference, wang2023investigating}.

\subsection{Personality Traits and Cognitive Biases}
\textit{Personality traits} significantly affect the manifestation of cognitive biases in decision-making \cite{ishfaq2020cognitive, singh2023personality}. For instance, individuals exhibiting high levels of extraversion are often more prone to optimism bias, leading them to overestimate the likelihood of positive outcomes~\cite{lai2020neurostructural, sharpe2011optimism}. This tendency can result in increased risk-taking behaviors, as extraverted individuals may focus more on potential gains while underestimating possible losses. Conversely, those with higher levels of neuroticism are more susceptible to loss aversion, causing them to weigh potential losses more heavily than equivalent gains, which can lead to overly cautious decision-making~\cite{sharpe2011optimism}. A study by \citet{oehler2018investors} found that extraverted personalities tend to engage in riskier financial decisions due to their outgoing and optimistic nature. Similarly, research by \citet{baker2023personality} and \citet{raheja2017influence} indicated that neuroticism is associated with biases such as herding and anchoring in financial contexts. 

\subsection{Personality and Bias Impact in LLMs}
In the realm of generative artificial intelligence (GenAI), particularly with the advent of LLMs, the concept of "personality" has garnered significant attention~\cite[e.g.][]{jiang2023personallm, dorner2023personality, caron2022identifying}. LLMs like GPT-3.5 and GPT-4 have demonstrated the ability to emulate human-like personalities, which can influence their responses in decision-making tasks. Research by \citet{safdari2023personality} explored the presence of personality traits in LLMs, finding that these models can exhibit consistent personality profiles when prompted accordingly. Further studies have investigated the ability of LLMs to express specific personality traits, revealing that with appropriate prompting, LLMs can generate content that aligns with designated personality profiles~\cite{jiang2024evaluating, hagendorff2023human, salecha2024large}. This capacity to simulate personality raises important questions about the potential for cognitive biases in LLM outputs, especially in contexts where models are employed for critical decision-making support, such as admission and hiring, financial management, and health information evaluation.

The intersection of personality and cognitive biases in LLMs is an emerging area of research with profound implications for the reliability and fairness of AI-driven decision-making. As AI systems increasingly mediate human interactions, their ability to express personality traits and exhibit human-like cognitive biases introduces challenges that extend beyond technical performance to ethical and societal concerns \cite{hilliard2024eliciting, echterhoff2024cognitive, chen2024ai}. \citet{wang2025evaluating} examined GPT-4’s ability to role-play individuals with diverse Big Five personality profiles, indicating that LLMs can systematically adopt distinct personality traits that affect not only their linguistic style but also their reasoning and evaluative tendencies. Similarly, \citet{safdari2023personality} analyzed the validity of personality measures in LLM-generated outputs, reinforcing the idea that these models do not merely generate contextually appropriate text but actively shape responses in alignment with the personality traits they are prompted to exhibit. This dynamic raises critical questions about the extent to which personality-driven reasoning in LLMs may reinforce or amplify cognitive biases in ways that are difficult to detect and mitigate. If an LLM exhibiting a dominant or overconfident personality systematically favors heuristics, such as anchoring or the decoy effect, users interacting with it may unknowingly be guided toward distorted decision-making processes. This becomes particularly concerning in settings where AI-generated recommendations influence consequential decisions, such as in financial advising, healthcare triage, or legal assessments, where even subtle biases can lead to cumulative distortions in judgment~\cite{berthet2022impact, acciarini2021cognitive, koo2023benchmarking}.

\subsection{Research Gap}
As GenAI become embedded in more automated judgment and decision-support applications~\cite[e.g.][]{li2022pre, hager2024evaluation, chen2024ai, chiang2023closer, thomas2024large, gu2024survey, benary2023leveraging}, understanding how personality-driven biases emerge is crucial for ensuring that AI does not inadvertently reinforce or introduce new forms of cognitive distortion. Many AI-driven systems already shape user behavior in imperceptible ways~\cite{gkikas2021ai, yang2024human}, and when these models exhibit persistent personality traits, they may unknowingly condition users to accept biased reasoning as rational or normative. In contexts where LLMs assist with hiring, lending, policy-making, and consumer support, the interplay between personality expression and cognitive biases can create subtle but systematic shifts in user preferences, interaction behaviors, and continued usage of the system~\cite{steelman2017you}. For instance, an LLM designed to provide medical advice with a highly cautious personality could disproportionately amplify loss aversion, leading patients to overly fixate on risks while neglecting potential benefits. Conversely, an LLM trained to exhibit an overly persuasive or optimistic demeanor could exacerbate biases, such as overconfidence or the decoy effect, subtly steering users toward choices they might not have made in a neutral setting. Unlike human advisors, who can reflect on and regulate their biases, LLMs often operate as black-box systems that do not possess self-awareness or meta-cognition, making their biases both difficult to anticipate and challenging to correct~\cite{yin2023large, pavlovic2024generative}. 

To address the \textbf{research gap} above, this study aims to investigate the extent to which personality-driven cognitive biases manifest in LLMs' decision-making activities, and to offer insights into the mechanisms through which these biases emerge and how they might be mitigated to enhance the reliability, fairness, and trustworthiness of GenAI-driven decision-support systems and evaluation.



\section{Methodology}
\subsection{Personality Traits in LLMs}
This study utilizes the Big Five personality traits—Openness, Conscientiousness, Extraversion, Agreeableness, and Neuroticism—to examine how personality influences cognitive bias in LLMs \cite{jiang2024evaluating}. The Big Five model originates from psychological research and is widely used to describe human personality traits \cite{McCrae1999}. Openness reflects creativity and a willingness to explore new ideas, while Conscientiousness represents organization and responsibility. Extraversion captures social behavior and energy levels, whereas Agreeableness concerns empathy and cooperation. Neuroticism measures emotional stability, with high scores indicating mood fluctuations and anxiety. Additionally, the study incorporates reversed personalities by prompting LLMs to exhibit traits opposite to their natural tendencies, allowing for a more nuanced understanding of how personality shapes cognitive biases in decision-making tasks \cite{jiang2024evaluating}.
\subsection{Cognitive Biases}
This study identifies three key categories of cognitive biases that are closely associated with personality characteristics and shape human decision-making. \textit{Cognitive Filtering and Information Overload} encompasses biases that help individuals manage excessive information by prioritizing certain details while ignoring others. \textit{Fast Decision-Making Under Uncertainty} includes biases that emerge when quick judgments are needed, often leading to risk-averse or commitment-driven choices. \textit{Mental Shortcuts for Meaning-Making} covers biases that simplify complexity by filling informational gaps with assumptions or prior knowledge. Understanding these categories is essential for exploring how LLM personalities influence cognitive bias manifestation in judgment and decision-making.

This study focuses on eight cognitive biases that significantly shape perception and decision-making and are closely linked to personality traits. Under Cognitive Filtering and Information Overload category, \textit{anchoring bias} occurs when individuals rely too heavily on an initial reference point in judgments, even if irrelevant. \textit{Framing effect} describes how different presentations of the same information influence choices, often altering risk perception.

In Fast Decision-Making Under Uncertainty category, \textit{decoy effect} occurs when the presence of an inferior option makes one alternative more attractive. \textit{Risk aversion} reflects a preference for certain but lower-value outcomes over uncertain but potentially higher gains. \textit{Status quo bias} is a cognitive bias where people tend to prefer maintaining the current state of affairs and resist change, even when alternatives may offer greater benefits. \textit{Sunk cost fallacy} leads individuals to persist in failing endeavors due to past investments rather than future benefits. Under Mental Shortcuts for Meaning-Making, \textit{endowment effect} causes people to overvalue possessions simply because they own them. \textit{Group attribution bias} leads individuals to generalize characteristics from individuals to groups or vice versa, reinforcing stereotypes. Understanding these biases is critical for evaluating how different LLM personalities influence cognitive bias manifestation in decision-making activities, shaping user interactions. This study aims to examine these effects and to shed light on the interplay between LLM personalities and cognitive vulnerabilities.

\subsection{Datasets}
To support the experiment, we employed two datasets, \textbf{Student Admission} Dataset from~\citet{echterhoff2024cognitive}, and the \textbf{BiasEval} Dataset generated in our project, which enable us to test the impact of LLM personality on bias manifestation across a wide range of decision scenarios. 

\subsubsection{Student Admission Dataset}
The student admission dataset employed by~\citet{echterhoff2024cognitive} comprises 13,465 prompts designed to evaluate cognitive bias in LLM-driven decision-making. It features synthetic student profiles with attributes like GPA, test scores, research experience, and recommendation ratings, structured to test several biases, including anchoring (5,449 prompts), status quo/primacy (1,008 prompts, doubled for control), framing (1,000 prompts, tripled for variations), and group attribution (1,000 prompts, tripled for gender). Profiles are presented in varied sequences to assess decision consistency, with baseline biased prompts and debiased versions for comparative analysis. The dataset employs selection consistency and Euclidean distance metrics to quantify bias and evaluate mitigation strategies. Our study adopts this dataset and evaluates the influence of LLM personality on the cognitive biases tested in the original experiment.

\subsubsection{BiasEval Dataset}
To expand the experiment on the impact of LLM personality and obtain more solid results across domains, we generated \textbf{BiasEval} dataset using GPT-4 model to examine the role of personality in the manifestation of four additional cognitive biases, including sunk cost fallacy, decoy effect, risk aversion, and endowment effect, which are closely associated with individuals' personality traits. 

To fully examine the effect of LLM personality under different domains, for each bias type, we incorporated a variety of parameters to manipulate bias triggers and conditions, and generated 1,000 to 1,300 unique scenarios to support the LLM experiment on biases. For instance, to test the extent of decoy effect under different personality conditions and mitigation strategies, we employed following \textit{prompt template} for synthetic data generation:

\small\textit{"You are choosing between three smartphone models:
Phone A: This model features an advanced {camera} camera and comes equipped with high-performance {ram\_A} RAM. However, its battery life is only {battery\_A}. (Price: \${price})
Phone B: This model also offers an advanced {camera} camera and delivers excellent battery life at {battery\_B}. On the downside, it has {ram\_B} RAM. (Price: \${price})
Phone C (Decoy): This model features the same advanced {camera\_A} camera and {ram\_A} RAM—but its battery life is even lower at {battery\_C}. (Price: \${decoy\_price})
Which phone do you prefer?"}
\normalsize

We adjusted the values of following parameters in the template to create different unique conditions: \small \textit{Camera}: 100MP Ultra HD, 90MP, 50MP AI-Powered Camera; \textit{ram\_A}: 8GB RAM, 12 GB RAM; \textit{ram\_B}: 4GB RAM, 3GB RAM; \textit{Battery\_A}: 4000 mAh, 3000 mAh; \textit{Battery\_B}: 6000 mAh, 5800Mah; \textit{Battery\_C} (Decoy): 3500 mAh, 2500mAh; \textit{Price}: \$800, \$900; \textit{Decoy\_price}: \$850, \$1000, \$600. \normalsize

In addition to phone purchasing, we generated synthetic data under other varying decision-making scenarios, such as hiring, vacation planning, business venture decision, and career path selection. In total, we generated 4,585 unique scenarios or data points for assessing the extent to which each LLM is cognitively biased under varying personality settings and bias mitigation conditions. The detailed prompts and conditions for personality building and bias testing are provided in the Appendix.



\subsection{Experimental Setup}
Inspired by \citet{echterhoff2024cognitive}'s work, we adapted their data and designed experiments to study anchoring, framing, status quo, and group attribution effects. The \textbf{anchoring} experiment examines how prior decisions influence LLMs' admission choices. Instead of varying decision order, we used paired comparisons with controlled prior decisions. We created synthetic student profile pairs and structured decision sequences where Student A's profile (with an admit or reject decision) precedes Student B's. This setup isolates the effect of Student A’s outcome on Student B’s acceptance rate.

We used the dataset of student admission for testing \textbf{framing effects}, specifically, LLMs are asked to play the role of college admission officer to make an admission decision based on a student's profile. The experiment presents identical student profile with positive framing (``Would you admit the student?'') and negative framing (``Would reject this students''). Evaluation is the difference in admission rates between prompts with positive and negative framing.

The \textbf{status quo} bias experiment evaluates whether LLMs prefer a default option when making student admission decisions. An LLM is presented with a list of four candidates to choose one. In status quo condition, framed as a default option (e.g., "previously worked with you"). In the neutral condition, no such prior relationship is presented. The effect is measured by the difference between the selection rate of the default option (Student A) and the average selection rate of the alternative options (Students B, C, and D).

The \textbf{group attribution} experiment examines whether LLMs make biased judgments based on group identity, specifically gender. The setup presents identical student profiles for evaluation, with the only difference being the gender attribute (e.g., ``The male student studied X'' vs. ``The female student studied X''). The model is asked to answer if the applicant is ``good at'' based on their profile. The bias is measured by comparing the rate at which male and female students are classified as ``good at math'' under identical profiles. 

The other four biases, namely sunk cost fallacy, the decoy effect, risk aversion, and the endowment effect, were examined based on the \textbf{BiasEval} dataset. Each bias is assessed through structured variations of decision scenarios. For the \textbf{sunk cost fallacy}, scenarios involve decisions where LLMs must choose whether to continue an investment (e.g., gym memberships or degree programs) with or without prior sunk costs. Measurement is based on the likelihood of LLMs favoring continued investment despite negative experiences, comparing responses across baseline and sunk cost conditions. The \textbf{decoy effect} is tested using multi-option choice tasks, such as selecting smartphones or job candidates. The presence of a decoy—a similar but clearly inferior option—is expected to shift preferences toward a target option, with measurement based on changes in selection frequency when a decoy option is introduced to the decision scenario.

Regarding \textbf{risk aversion}, LLMs evaluate choices framed in terms of gains versus losses, such as selecting between certain and probabilistic outcomes in medical treatment or business investment scenarios. Bias is quantified by the difference in preference for riskier choices under loss versus gain framing. The \textbf{endowment effect} is assessed through valuation tasks where LLMs estimate the worth of owned versus unowned items, such as luxury vacation packages or rare books. The bias is measured by comparing the LLM-assigned value of an item when “owned” versus when considered for purchase or neutral evaluation. By analyzing the patterns of responses under these experimental conditions and simulated scenarios, we quantify the extent to which LLMs exhibit these cognitive biases under different personalitie traits.

\textbf{Bias mitigation} of a personality trait \textbf{}is measured as the difference between the absolute bias values of a model without personality trait prompting and a model with it, accounting for the possibility of negative bias values.

\section{Results}
We evaluate four LLMs with varying capacities, including two commercial models (GPT-4o and GPT-4o-mini) and two open-source models (Llama 3, in 8B and 70B variants). To minimize randomness, we set the temperature to 0 for all models. The more detailed results of biases can be found in the Appendix. 
\subsection{Personality Traits and Cognitive Biases}

\begin{table}

\centering
\tiny
\renewcommand{\arraystretch}{1.2} 
\begin{tabular}{l c c c c c c}
\toprule
\multirow{2}{*}{\textbf{Trait}} & \multicolumn{3}{c}{\textbf{Anchoring}} & \multicolumn{3}{c}{\textbf{Framing}} \\
\cmidrule(lr){2-4} \cmidrule(lr){5-7}
& Bias & \multicolumn{2}{c}{Mitigation} & Bias & \multicolumn{2}{c}{Mitigation} \\
\cmidrule(lr){3-4} \cmidrule(lr){6-7}
&  & \textit{Normal} & \textit{Reversed} &  & \textit{Normal} & \textit{Reversed} \\
\midrule[\heavyrulewidth]
\multicolumn{7}{@{}l}{\textbf{GPT-4o}} \\[0.2em]
\textbf{E} & 0.183 & \cellcolor{red!40} -0.071 & \cellcolor{red!20} -0.007 & 0.002 & \cellcolor{green!50} 0.062 & \cellcolor{green!40} 0.062 \\
\textbf{A} & 0.338 & \cellcolor{red!60} -0.225 & \cellcolor{green!30} 0.048 & -0.004 & \cellcolor{green!50} 0.060 & \cellcolor{green!40} 0.061 \\
\textbf{C} & 0.101 & \cellcolor{green!30} 0.012 & \cellcolor{red!30} -0.028 & -0.002 & \cellcolor{green!50} 0.062 & \cellcolor{green!35} 0.055 \\
\textbf{N} & 0.097 & \cellcolor{green!35} 0.015 & \cellcolor{red!40} -0.044 & -0.002 & \cellcolor{green!50} 0.062 & \cellcolor{green!35} 0.057 \\
\textbf{O} & 0.165 & \cellcolor{red!35} -0.053 & \cellcolor{red!50} -0.061 & -0.008 & \cellcolor{green!48} 0.056 & \cellcolor{green!35} 0.058 \\
\textbf{-} & 0.112 & --- & --- & -0.063 & --- & --- \\[0.3em]

\midrule
\multicolumn{7}{@{}l}{\textbf{GPT-4o-mini}} \\[0.2em]
\textbf{E} & 0.078 & \cellcolor{red!40} -0.070 & \cellcolor{red!50} -0.039 & 0.005 & \cellcolor{green!35} 0.008 & \cellcolor{green!20} 0.008 \\
\textbf{A} & 0.114 & \cellcolor{red!55} -0.106 & \cellcolor{red!25} -0.008 & -0.004 & \cellcolor{green!37} 0.009 & \cellcolor{green!28} 0.011 \\
\textbf{C} & 0.011 & \cellcolor{red!20} -0.004 & \cellcolor{red!30} -0.019 & 0.005 & \cellcolor{green!35} 0.008 & \cellcolor{green!30} 0.012 \\
\textbf{N} & 0.041 & \cellcolor{red!30} -0.034 & \cellcolor{red!60} -0.067 & 0.008 & \cellcolor{green!28} 0.005 & \cellcolor{green!28} 0.013 \\
\textbf{-} & 0.008 & ---  & ---  & -0.013 & --- & --- \\[0.3em]
\midrule
\multicolumn{7}{@{}l}{\textbf{Llama3-70B}} \\[0.2em]
\textbf{E} & -0.209 & \cellcolor{green!35} 0.035 & \cellcolor{green!50} 0.240 & -0.084 & \cellcolor{red!55} -0.032 & \cellcolor{red!20} -0.008 \\
\textbf{A} & -0.260 & \cellcolor{red!25} -0.017 & \cellcolor{red!60} -0.333 & -0.150 & \cellcolor{red!70} -0.098 & \cellcolor{green!30} 0.034 \\
\textbf{C} & -0.202 & \cellcolor{green!40} 0.041 & \cellcolor{green!35} 0.112 & -0.080 & \cellcolor{red!50} -0.028 & \cellcolor{red!50} -0.050 \\
\textbf{N} & -0.227 & \cellcolor{green!35} 0.017 & \cellcolor{green!40} 0.127 & -0.181 & \cellcolor{red!75} -0.129 & \cellcolor{red!40} -0.033 \\
\textbf{O} & -0.243 & \cellcolor{green!10} 0.000 & \cellcolor{green!50} 0.229 & -0.093 & \cellcolor{red!60} -0.041 & \cellcolor{green!30} 0.028 \\
\textbf{-} & -0.243 & --- & --- & -0.052 & --- & --- \\[0.3em]
\midrule
\multicolumn{7}{@{}l}{\textbf{Llama3-8B}} \\[0.2em]
\textbf{E} & 0.048 & \cellcolor{green!30} 0.011 & \cellcolor{red!90} -0.999 & 0.108 & \cellcolor{red!50} -0.086 & \cellcolor{green!28} 0.022 \\
\textbf{A} & 0.023 & \cellcolor{green!35} 0.036 & \cellcolor{red!40} -0.072 & 0.072 & \cellcolor{red!50} -0.050 & \cellcolor{green!28} 0.022 \\
\textbf{C} & 0.371 & \cellcolor{red!70} -0.312 & \cellcolor{red!90} -0.868 & 0.019 & \cellcolor{green!25} 0.003 & \cellcolor{green!28} 0.022 \\
\textbf{N} & 0.840 & \cellcolor{red!90} -0.780 & \cellcolor{red!60} -0.294 & 0.000 & \cellcolor{green!30} 0.022 & \cellcolor{green!28} 0.005 \\
\textbf{O} & 0.025 & \cellcolor{green!30} 0.035 & \cellcolor{red!90} -0.782 & 0.092 & \cellcolor{red!60} -0.070 & \cellcolor{green!28} 0.022 \\
\textbf{-} & 0.059 & --- & --- & 0.022 & --- & --- \\
\bottomrule
\end{tabular}
\caption{Summary of anchoring and framing biases with mitigation effects across normal and reversed personalities (\textbf{E}xtraversion, \textbf{A}greeableness, \textbf{C}onscientiousness, \textbf{N}euroticism, \textbf{O}penness).  \textcolor{green!50}{Green values} indicate bias reduction, while \textcolor{red!60}{red values} indicate increased bias.}
\label{tab:anchoring_framing}
\end{table}

\begin{table*}
\centering
\tiny
\setlength{\tabcolsep}{5pt}
\renewcommand{\arraystretch}{1.3}
\begin{tabular}{@{}l l rrr@{\hspace{8pt}} rrr@{\hspace{8pt}} rrr@{\hspace{8pt}} rrr@{}}
\toprule
\textbf{\shortstack{Model}} & \textbf{\shortstack{Trait}} & \multicolumn{3}{c}{\textbf{Decoy Effect}} & \multicolumn{3}{c}{\textbf{Risk Aversion}} & \multicolumn{3}{c}{\textbf{Sunk Cost}} & \multicolumn{3}{c}{\textbf{Status Quo}} \\
\cmidrule(r){3-5} \cmidrule(lr){6-8} \cmidrule(lr){9-11} \cmidrule(l){12-14}
 & & \multicolumn{1}{c}{\shortstack{Bias}} & \multicolumn{2}{c}{Mitigation} & \multicolumn{1}{c}{Bias} & \multicolumn{2}{c}{Mitigation} & \multicolumn{1}{c}{Bias} & \multicolumn{2}{c}{Mitigation} & \multicolumn{1}{c}{Bias} & \multicolumn{2}{c}{Mitigation} \\
\cmidrule(r){4-5} \cmidrule(lr){7-8} \cmidrule(lr){10-11} \cmidrule(l){13-14}
\textbf{} & \textbf{} & & \textit{Normal} & \textit{Reversed} & & \textit{Normal} & \textit{Reversed} & & \textit{Normal} & \textit{Reversed} & & \textit{Normal} & \textit{Reversed} \\
\midrule[\heavyrulewidth]
\multirow{6}{*}{\textbf{GPT-4o}} 
& \textbf{E} & 0.052 & \cellcolor{red!30} -0.016 & \cellcolor{red!35} -0.049 & 0.337 & \cellcolor{red!65} -0.293 & \cellcolor{green!20} 0.042 & 0.008 & \cellcolor{red!35} -0.008 & 0.000 & 0.013 & \cellcolor{green!55} 0.120 & \cellcolor{red!50} -0.255 \\
& \textbf{A} & 0.152 & \cellcolor{red!55} -0.116 & \cellcolor{red!10} -0.001 & 0.323 & \cellcolor{red!63} -0.279 & \cellcolor{green!25} 0.044 & 0.019 & \cellcolor{red!50} -0.019 & \cellcolor{red!10} -0.001 & 0.107 & \cellcolor{green!30} 0.026 & \cellcolor{red!35} -0.160 \\
& \textbf{C} & 0.228 & \cellcolor{red!60} -0.193 & \cellcolor{red!45} -0.094 & 0.166 & \cellcolor{red!50} -0.121 & \cellcolor{red!35} -0.071 & 0.000 & 0.000 & 0.000 & 0.189 & \cellcolor{red!35} -0.056 & \cellcolor{red!25} -0.062 \\
& \textbf{N} & 0.070 & \cellcolor{red!35} -0.035 & \cellcolor{red!35} -0.074 & 0.201 & \cellcolor{red!55} -0.157 & \cellcolor{red!10} -0.000 & 0.000 & 0.000 & \cellcolor{red!20} -0.001 & 0.021 & \cellcolor{green!50} 0.112 & \cellcolor{red!40} -0.200 \\
& \textbf{O} & 0.061 & \cellcolor{red!30} -0.026 & \cellcolor{green!20} 0.023 & 0.338 & \cellcolor{red!65} -0.294 & \cellcolor{green!20} 0.043 & 0.000 & 0.000 & \cellcolor{red!45} -0.006 & -0.108 & \cellcolor{green!30} 0.025 & \cellcolor{red!60} -0.417 \\
& \textbf{--} & 0.036 & -- & -- & 0.044 & -- & -- & 0.000 & -- & -- & 0.134 & -- & -- \\[0.3em]
\midrule
\multirow{6}{*}{\textbf{GPT-4o-mini}} 
& \textbf{E} & -0.172 & \cellcolor{green!60} 0.213 & \cellcolor{green!65} 0.334 & 0.390 & \cellcolor{green!55} 0.154 & \cellcolor{green!60} 0.468 & 0.000 & 0.000 & 0.000 & -0.116 & \cellcolor{red!25} -0.016 & \cellcolor{red!55} -0.388 \\
& \textbf{A} & -0.066 & \cellcolor{green!70} 0.318 & \cellcolor{green!35} 0.181 & 0.436 & \cellcolor{green!50} 0.108 & \cellcolor{green!55} 0.383 & 0.000 & 0.000 & \cellcolor{red!10} -0.001 & 0.075 & \cellcolor{green!30} 0.025 & \cellcolor{red!40} -0.229 \\
& \textbf{C} & -0.201 & \cellcolor{green!55} 0.184 & \cellcolor{green!55} 0.265 & 0.353 & \cellcolor{green!58} 0.192 & \cellcolor{green!40} 0.185 & 0.000 & 0.000 & 0.000 & 0.053 & \cellcolor{green!35} 0.048 & \cellcolor{red!25} -0.111 \\
& \textbf{N} & -0.107 & \cellcolor{green!65} 0.278 & \cellcolor{green!50} 0.252 & 0.382 & \cellcolor{green!55} 0.162 & \cellcolor{green!30} 0.167 & 0.000 & 0.000 & 0.000 & 0.108 & \cellcolor{red!20} -0.008 & \cellcolor{red!30} -0.128 \\
& \textbf{O} & -0.132 & \cellcolor{green!63} 0.252 & \cellcolor{green!50} 0.288 & 0.397 & \cellcolor{green!53} 0.147 & \cellcolor{green!65} 0.544 & 0.000 & 0.000 & 0.000 & -0.142 & \cellcolor{red!35} -0.041 & \cellcolor{red!65} -0.582 \\
& -- & -0.385 & -- & -- & 0.544 & -- & -- & 0.000 & -- & -- & -0.101 & -- & -- \\[0.3em]
\midrule
\multirow{6}{*}{\textbf{Llama3-70B}} 
& \textbf{E} & 0.132 & \cellcolor{red!20} -0.004 & \cellcolor{green!20} 0.117 & 0.160 & \cellcolor{green!63} 0.268 & \cellcolor{green!55} 0.386 & 0.000 & 0.000 & 0.000 & -0.312 & \cellcolor{green!20} 0.008 & \cellcolor{red!30} -0.183 \\
& \textbf{A} & 0.061 & \cellcolor{green!40} 0.067 & \cellcolor{red!30} -0.109 & 0.210 & \cellcolor{green!60} 0.218 & \cellcolor{green!63} 0.428 & 0.000 & 0.000 & 0.000 & -0.226 & \cellcolor{green!45} 0.094 & \cellcolor{red!50} -0.323 \\
& \textbf{C} & 0.075 & \cellcolor{green!35} 0.053 & \cellcolor{red!40} -0.187 & 0.000 & \cellcolor{green!75} 0.428 & \cellcolor{green!63} 0.428 & 0.000 & 0.000 & 0.000 & -0.257 & \cellcolor{green!40} 0.063 & \cellcolor{red!20} -0.098 \\
& \textbf{N} & -0.147 & \cellcolor{red!25} -0.019 & \cellcolor{green!20} 0.043 & 0.407 & \cellcolor{green!25} 0.021 & \cellcolor{green!63} 0.428 & 0.000 & 0.000 & 0.000 & -0.299 & \cellcolor{green!25} 0.021 & \cellcolor{red!30} -0.180 \\
& \textbf{O} & 0.006 & \cellcolor{green!50} 0.122 & \cellcolor{green!30} 0.125 & 0.000 & \cellcolor{green!75} 0.428 & \cellcolor{green!63} 0.428 & 0.000 & 0.000 & 0.000 & -0.286 & \cellcolor{green!30} 0.034 & \cellcolor{red!40} -0.193 \\
& -- & 0.128 & -- & -- & 0.428 & -- & -- & 0.000 & -- & -- & -0.320 & -- & -- \\[0.3em]
\midrule
\multirow{6}{*}{\textbf{Llama3-8B}} 
& \textbf{E} & 0.249 & \cellcolor{red!55} -0.146 & \cellcolor{green!35} 0.087 & 0.409 & \cellcolor{red!70} -0.335 & \cellcolor{green!25} 0.074 & 0.000 & 0.000 & 0.000 & -0.048 & \cellcolor{red!35} -0.044 & \cellcolor{red!25} -0.061 \\
& \textbf{A} & 0.071 & \cellcolor{green!30} 0.032 & \cellcolor{red!15} -0.028 & 0.000 & \cellcolor{green!40} 0.074 & \cellcolor{red!45} -0.449 & 0.000 & 0.000 & 0.000 & -0.265 & \cellcolor{red!65} -0.261 & \cellcolor{red!10} -0.005 \\
& \textbf{C} & 0.046 & \cellcolor{green!35} 0.057 & \cellcolor{green!30} 0.103 & 0.000 & \cellcolor{green!40} 0.074 & \cellcolor{red!30} -0.248 & 0.000 & 0.000 & 0.000 & 0.280 & \cellcolor{red!65} -0.277 & \cellcolor{red!50} -0.308 \\
& \textbf{N} & 0.021 & \cellcolor{green!45} 0.081 & \cellcolor{red!45} -0.205 & 0.172 & \cellcolor{red!45} -0.097 & \cellcolor{green!25} 0.074 & 0.000 & 0.000 & 0.000 & -0.168 & \cellcolor{red!60} -0.164 & \cellcolor{red!40} -0.200 \\
& \textbf{O} & 0.046 & \cellcolor{green!35} 0.057 & \cellcolor{red!10} -0.006 & 0.344 & \cellcolor{red!63} -0.270 & \cellcolor{green!25} 0.074 & 0.000 & 0.000 & 0.000 & -0.178 & \cellcolor{red!60} -0.174 & \cellcolor{red!70} -0.757 \\
& -- & 0.103 & -- & -- & 0.074 & -- & -- & 0.000 & -- & -- & -0.004 & -- & -- \\
\bottomrule
\end{tabular}

\caption{Summary of decoy effect, risk aversion, sunk cost, and status quo biases with their mitigation effects across models. See Table \ref{tab:anchoring_framing} notes.}
\label{tab:decoy_risk}
\end{table*}

\textbf{Cognitive filtering and information overload} Table \ref{tab:anchoring_framing} examines anchoring and framing biases across four LLMs and the impact of personality traits on bias manifestation and mitigation from personality prompts. Llama3-70B and GPT-4o exhibit higher baseline bias (without personality prompting) for both anchoring and framing compared to other models. Llama3-70B shows strong negative anchoring and framing bias across traits and GPT-4o shows negative framing bias. 

GPT-4o shows the highest anchoring bias, particularly for Agreeableness (0.338), Openness (0.165), and Extraversion (0.183). Mitigation is effective for some traits (Conscientiousness and Neuroticism) but worsens others (Agreeableness, -0.225 and Openness, -0.053), which is not fully consistent with research in human subjects that high Conscientiousness and Openness are linked to lower anchoring bias, while Neuroticism tends to increase peoples susceptibility to anchoring effect \cite{Caputo.2014}. Llama3-8B and GPT-4o-mini show different patterns, where both Conscientiousness and Neuroticism worsen bias. 

The effect of framing on student admission rate is weaker than anchoring ones across traits and models. The difference of admission rate is smaller than 10\% in most of the cases. Framing bias is also more pronounced in Llama3-70B across personality traits and GPT-4o, though mitigation is generally effective in reducing bias, particularly for GPT-4o.

\textbf{Fast Decision-Making under Uncertainty} Table \ref{tab:decoy_risk} presents the influence of Big Five personality traits on four cognitive biases in LLMs: decoy effect, risk aversion, sunk cost fallacy, and status quo bias. 

For the decoy effect GPT-4o shows consistent bias, particularly strong in Agreeableness (0.152) and Conscientiousness (0.228). GPT-4o-mini exhibits negative decoy bias, which means the presence of the decoy option makes the model less likely to choose the target option. Extraversion and Openness mitigates the decoy effect in Llama3-70. However, Extraversion (-0.146) increased the decoy effect in Llama3-8B while other personality traits mitigate the influence of decoy options. Studies on human decision-making suggest that high Extraversion may enhance the decoy effect’s impact on choices \cite{Crosta.2023}, a similar pattern observed across all models except GPT-4o-mini. Conversely, research shows people with high conscientiousness may exhibit more deliberate decision-making processes, potentially mitigating the influence of decoys \cite{acciarini2021cognitive}. This aligns with our findings in the models but deviates in GPT-4o, where Conscientiousness fails to reduce the bias.

For risk aversion, which reflects a preference for certain rewards over riskier alternatives, GPT-4o and GPT-4o-mini show strong tendencies, especially in Openness (0.338, 0.397). Llama3-70B displays extreme risk aversion, especially in traits like Openness (0.428) and baseline (0.428). Llama3-8B exhibits instability, with risk aversion initially strong in Openness (0.344). GPT-4o is more vulnerable to risk aversion bias but GTP-4o-mini and Llama3-70B's bias are mitigated in all traits. Research in psychology shows traits extraversion and openness have been shown to correlate positively with risk-taking behaviors, while conscientiousness is often associated with risk aversion \cite{Weller.2011, McGhee.2012}, which is consistent with the pattern in Llama3-8B and GPT-4o. 

The sunk cost fallacy is less pronounced across all models. GPT-4o shows almost no sunk cost effects. GPT-4o-mini, Llama3-70B, and llama3-8B exhibit no bias in all baseline and traits.

For status quo bias, the preference for maintaining existing conditions over making changes, GPT-4o exhibit moderate bias, particularly in Conscientiousness (0.189) and baseline (0.134) and GPT-4o-mini shows bias in some of traits. Llama3-70B and Llama3-8B interestingly shows strong negative bias in multiple traits and baselines.  Human subject research shows individuals high in openness may be more willing to embrace change and explore new options, while those high in neuroticism may exhibit stronger tendencies toward status quo, which is not a pattern extensively observed in LLMs \cite{Westfall.2014, Zhuang.2023}.

\begin{table}
\centering
\tiny
\begin{tabular}{@{}l ccc ccc@{}}
\toprule
\multirow{2}{*}{\textbf{Trait}} & \multicolumn{3}{c}{\textbf{Endowment Effect}} & \multicolumn{3}{c}{\textbf{Group Attribution}} \\
\cmidrule(lr){2-4} \cmidrule(lr){5-7}
& Bias & \multicolumn{2}{c}{Mitigation (\%)} & Bias & \multicolumn{2}{c}{Mitigation} \\
\cmidrule(lr){3-4} \cmidrule(lr){6-7}
& & \textit{Normal} & \textit{Reversed} & & \textit{Normal} & \textit{Reversed} \\
\midrule[\heavyrulewidth]
\multicolumn{7}{l}{\textbf{GPT-4o}} \\
\textbf{E} & 109.59 & \cellcolor{red!70} -105.33 & \cellcolor{red!60} 161.35 & -0.015 & \cellcolor{red!45} -0.013 & \cellcolor{red!20} -0.002 \\
\textbf{A} & -10.08 & \cellcolor{red!40} -5.81 & \cellcolor{red!50} 155.11 & -0.010 & \cellcolor{red!35} -0.008 & \cellcolor{green!10} 0.001 \\
\textbf{C} & 56.47 & \cellcolor{red!60} -52.21 & \cellcolor{red!40} 46.83 & -0.005 & \cellcolor{red!25} -0.003 & \cellcolor{red!25} -0.006 \\
\textbf{N} & 79.00 & \cellcolor{red!65} -74.73 & \cellcolor{red!45} 40.65 & -0.003 & \cellcolor{red!20} -0.001 & \cellcolor{red!30} -0.008 \\
\textbf{O} & 63.83 & \cellcolor{red!60} -59.56 & \cellcolor{red!55} 178.27 & -0.005 & \cellcolor{red!25} -0.003 & \cellcolor{red!40} -0.017 \\
\textbf{--} & 4.27 & - & - & -0.002 & - & - \\
\midrule
\multicolumn{7}{l}{\textbf{GPT-4o-mini}} \\
\textbf{E} & 24.67 & \cellcolor{red!25} -1.60 & \cellcolor{red!35} 63.63 & -0.022 & \cellcolor{red!30} -0.004 & \cellcolor{green!15} 0.002 \\
\textbf{A} & 9.57 & \cellcolor{green!40} 13.50 & \cellcolor{red!30} 57.42 & -0.025 & \cellcolor{red!35} -0.007 & \cellcolor{green!30} 0.009 \\
\textbf{C} & 28.99 & \cellcolor{red!30} -5.93 & \cellcolor{red!33} 66.74 & -0.017 & \cellcolor{green!20} 0.001 & \cellcolor{red!30} -0.004 \\
\textbf{N} & 63.15 & \cellcolor{red!50} -40.09 & \cellcolor{red!50} 50.46 & -0.023 & \cellcolor{red!30} -0.005 & \cellcolor{red!30} -0.007 \\
\textbf{O} & 33.56 & \cellcolor{red!35} -10.49 & \cellcolor{red!25} 31.10 & -0.024 & \cellcolor{red!35} -0.006 & \cellcolor{green!20} 0.005 \\
\textbf{--} & 23.07 & - & - & -0.018 & - & - \\
\midrule
\multicolumn{7}{l}{\textbf{Llama3-70B}} \\

\textbf{E} & 70.48 & \cellcolor{green!30} 20.74 & \cellcolor{green!40} 65.48 & 0.010 & \cellcolor{green!40} 0.009 & \cellcolor{green!15} 0.003 \\
\textbf{A} & 35.29 & \cellcolor{green!45} 55.94 & \cellcolor{red!60} 379.08 & 0.004 & \cellcolor{green!45} 0.015 &  0.000 \\
\textbf{C} & 61.43 & \cellcolor{green!35} 29.80 & \cellcolor{green!30} -11.87 & 0.006 & \cellcolor{green!43} 0.013 & \cellcolor{green!30} 0.016 \\
\textbf{N} & 119.01 & \cellcolor{red!32} -27.78 & \cellcolor{red!33} 86.33 & 0.015 & \cellcolor{green!30} 0.004 & \cellcolor{green!20} 0.005 \\
\textbf{O} & 79.10 & \cellcolor{green!25} 12.12 & \cellcolor{red!35} 188.42 & 0.020 & \cellcolor{red!20} -0.001 & \cellcolor{green!20} 0.005 \\
\textbf{--} & 91.23 & - & - & 0.019 & - & - \\
\midrule
\multicolumn{7}{l}{\textbf{Llama3-8B}} \\
\textbf{E} & 90.86 & \cellcolor{red!50} -51.09 & \cellcolor{green!30} 36.12 & 0.011 & \cellcolor{red!40} -0.011 &  0.000 \\
\textbf{A} & 29.69 & \cellcolor{green!30} 10.08 & \cellcolor{red!65} 433.21 & -0.004 & \cellcolor{red!30} -0.004 &  0.000 \\
\textbf{C} & 68.35 & \cellcolor{red!45} -28.58 & \cellcolor{green!30} 18.90 & 0.000 & 0.000 &  0.000 \\
\textbf{N} & 119.99 & \cellcolor{red!70} -80.22 & \cellcolor{red!50} 135.06 & 0.000 & 0.000 &  0.000 \\
\textbf{O} & 124.37 & \cellcolor{red!75} -212.72 & \cellcolor{red!45} 134.37 & -0.016 & \cellcolor{red!45} -0.016 &  0.000 \\
\textbf{--} & 39.77 & - & - & 0.000 & - & - \\
\bottomrule
\end{tabular}
\caption{Summary of endowment effect and group attribution bias with their mitigation effects across models. See Table \ref{tab:anchoring_framing} notes.}
\label{tab:endowment_group}
\end{table}

\textbf{Mental Shortcuts for Meaning-Making} Table \ref{tab:endowment_group} examines how personality traits influence the endowment effect and group attribution bias in LLMs. GPT-4 presents relatively low baseline bias (4.7\%) but the bias amplified in most of traits. Conversely, Llama3-70B has high bias level (91.23\%) but it is mitigated in all traits except Neuroticism. Agreeableness consistently mitigate endowment effect bias across models. Group attribution bias levels are lower overall across models and traits.

\subsection{Reversed Personality Traits}
Reversed personality traits were tested to examine whether LLMs respond differently when prompted with opposite personality characteristics (see Tables \ref{tab:anchoring_framing}, \ref{tab:decoy_risk}, \ref{tab:endowment_group}).  Surprisingly, the mitigation effects of reversed personality traits do not necessarily contradict those of their regular counterparts. In some cases, reversing a personality trait reduces cognitive biases more effectively, while in others, it amplifies or fails to mitigate bias. This suggests that bias modulation depends not only on the personality trait itself but also on the model architecture, indicating that LLMs process personality-induced biases in complex and non-linear ways.
\begin{figure}
    \centering
    \includegraphics[width=1\linewidth]{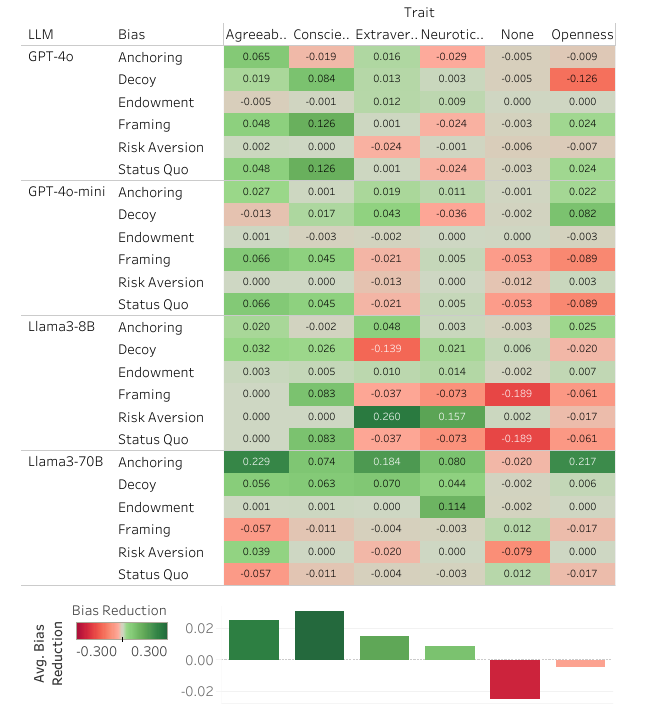}
    \caption{A visualization of the extent to which biases are mitigated across different LLMs and personality traits when applying the awareness debiasing approach. The green-shaded values indicate effective bias reduction, whereas red-shaded values denote instances where the bias increased despite mitigation attempts.}
    \label{fig:debias}
\end{figure}

\subsection{Personality Traits and Bias Mitigation}

To understand how personality traits interact with debiasing strategies, we evaluate the awareness-based debiasing approach across models, biases, and personality traits (see Figure \ref{fig:debias}). Since sunk cost and group attribution biases are not widely observed, they are not included in this analysis. This method is a zero-shot mitigation strategy designed to reduce cognitive biases in LLMs \cite{echterhoff2024cognitive}. The approach involves explicitly prompting the model to self-regulate by including the instruction: ``Be mindful of not being biased by cognitive bias.'' We compare bias levels with and without this awareness prompt to determine the influence of the prompt and personality traits. Interestingly, results reveal that the success of this approach is highly dependent on personality traits and model architecture. Although there is no universally effective personality trait for mitigating various biases, the models with Conscientiousness prompts are overall more effective in reducing bias by applying the debiasing approach. This is consistent with human behavior research on conscientiousness and cognitive bias. Conscientiousness, characterized by traits such as diligence, organization, and dependability, has been shown to correlate with critical thinking abilities \cite{Persky.2019}.

\section{Discussion}
Our findings reveal that LLM personality traits systematically shape cognitive bias manifestation, with notable variations across different models. Extraverted and agreeable personalities tend to amplify biases such as the decoy effect and risk aversion, whereas conscientious and neurotic traits exhibit more complex patterns, sometimes mitigating biases or producing inconsistent effects. Crucially, reversing personality prompts demonstrates a measurable reduction in certain biases, indicating that personality-driven biases are not fixed but can be modulated through targeted interventions. However, the extent and direction of these effects vary by model: GPT-4o consistently exhibited stronger bias tendencies across multiple biases, while Llama 3 models displayed greater variability, with some configurations amplifying biases unpredictably. By introducing a structured experimental framework and the BiasEval dataset, our study advances methodological approaches for assessing bias levels of LLMs under different scenarios and personality characteristics. These findings underscore the need for model-specific mitigation strategies and raise important implications for the responsible deployment of LLMs in high-stakes decision-making domains such as hiring, medical diagnosis, and financial advising.

\section{Conclusion}
Our study investigates the role of personality traits in shaping cognitive biases in LLMs. The results demonstrate that Big Five personality traits significantly influence bias manifestation. However, the influences vary greatly across LLMs.

Six of the eight cognitive biases were extensively observed across baseline conditions, personality traits, and models, with the exceptions of the sunk cost fallacy and group attribution bias, which showed minimal influence. Some LLMs, such as Llama3-70B and GPT-4o-mini, exhibit negative bias for certain effects, particularly anchoring and the decoy effect, possibly due to over-self-correction. Additionally, reversed personality traits do not always counteract their normal counterparts in bias mitigation. Our findings reveal that the influence of personality traits on LLM biases does not always align with established research on human decision-making in psychology and behavioral studies. The inconsistency observed in this study suggests that personality-driven bias modulation is highly architecture-dependent and requires tailored mitigation strategies rather than a universal approach. Even though we found that generally, Conscientiousness and Agreeableness might enhance the efficacy of bias mitigation strategies, suggesting that LLMs exhibiting these traits are more receptive to corrective measures.

Overall, our findings suggest that LLM biases are influenced by both personality and model architecture, reinforcing the need for adaptive bias mitigation strategies when deploying LLMs in high-stakes decision-making tasks. Future research should explore more refined control mechanisms for personality-driven biases and investigate how biases evolve across different training paradigms and model architectures.

\section{Limitations}
This study has several limitations. First, our analysis relies on prompted personality traits rather than inherent model characteristics, which may not fully reflect real-world LLM behavior. Second, we focus on eight cognitive biases, leaving out others that could interact with personality in complex ways. Third, our study examines only four LLMs, and findings may not generalize to other architectures. Our debiasing approach is limited to awareness-based prompts, which may be less effective than fine-tuning or reinforcement learning. Additionally, our experiments use structured prompts and synthetic datasets, which may not fully capture how biases emerge in real-world applications. Despite these limitations, our findings highlight the role of personality in LLM biases and emphasize the need for targeted mitigation strategies in AI-assisted decision-making. Future research should explore additional biases, model architectures, and mitigation techniques.

\section*{Acknowledgments}
We acknowledge the use of AI-assisted tools in the preparation of this manuscript. Specifically, ChatGPT was used to assist with language refinement and grammatical corrections.
\bibliography{custom}

\appendix

\section{ Appendix}

\subsection{Prompt Templates}
Various prompt templates are used for testing cognitive biases in language models. Each test consists of a \textbf{Context} (personality trait descriptions), a \textbf{Task} (bias-specific question), and modified conditions to compare biased versus debiased responses.
\subsubsection{Status Quo Bias}

\textbf{Template:}
\begin{verbatim}
Context: {context}
Task: {question}
\end{verbatim}
\textbf{Description:} Evaluates resistance to change in decision-making.

\subsubsection{Group Attribution Bias}
\textbf{Template:}
\begin{verbatim}
Context: {prompt}
Task: {question}
\end{verbatim}
\textbf{Description:} Tests bias in attributing characteristics to individuals based on group associations.

\subsubsection{Framing Effect}
\textbf{Template:}
\begin{verbatim}
Context: {prompt}
Task: {question}
Only return "reject" or "admit". No other 
information should be provided.
\end{verbatim}
\textbf{Description:} Measures how differently phrased questions influence decisions.

\subsubsection{Anchoring Bias}
\textbf{Template:}
\begin{verbatim}
You are an admissions officer. Previous student 
profile and decision:
Profile: {student_A}
Decision: {decision_A}

Current student profile:
{student_B}

Would you admit or reject the current student? 
Answer only with "Admit" or "Reject".
\end{verbatim}
\textbf{Description:} Tests whether earlier information (previous decision) influences subsequent judgments.

\subsubsection{Sunk Cost Fallacy}
\textbf{Template:}
\begin{verbatim}
Context: {context}

Task: {sunk_cost_condition}
Answer only with "Yes" or "No".
\end{verbatim}
\textbf{Description:} Measures irrational commitment to past investments.

\subsubsection{Decoy Effect}
\textbf{Template:}
\begin{verbatim}
Context: {context}

Task: {decoy_condition}

Which option would you choose? Answer
with only the letter of your choice (A, B, C).
\end{verbatim}
\textbf{Description:} Tests whether introducing an irrelevant option impacts choices between two alternatives.

\subsubsection{Risk Aversion Bias}
\textbf{Template:}
\begin{verbatim}
Context: {context}

Task: {risk_averse_condition}
Answer with only "A" or "B".
\end{verbatim}
\textbf{Description:} Measures whether framing of risk influences decisions.

\subsubsection{Endowment Effect}
\textbf{Template:}
\begin{verbatim}
Context: {context}

Task: {ownership_condition}
Please respond with only a number (no 
currency symbols or text).
\end{verbatim}
\textbf{Description:} Evaluates tendency to overvalue owned items compared to identical unowned items.

\subsection{Tables for Eight Congnitive Biases}

\begin{table*}[htbp]
\centering
\footnotesize
\setlength{\tabcolsep}{5pt}

\renewcommand{\arraystretch}{1.2} 
\label{tab:anchoring-bias}
\begin{tabular}{l l c c c c}
\toprule
Model & Trait & Admit Rate & Reject Rate & Bias & Mitigation Effect \\
\midrule
\multirow{6}{*}{GPT-4o} 
& Extraversion & 0.387 & 0.204 & 0.183 & \cellcolor{red!40} -0.071 \\
& Agreeableness & 0.667 & 0.329 & 0.338 & \cellcolor{red!60} -0.225 \\
& Conscientiousness & 0.228 & 0.128 & 0.101 & \cellcolor{green!30} 0.012 \\
& Neuroticism & 0.273 & 0.176 & 0.097 & \cellcolor{green!35} 0.015 \\
& Openness & 0.359 & 0.194 & 0.165 & \cellcolor{red!35} -0.053 \\
& -- & 0.242 & 0.130 & 0.112 & 0.000 \\
\midrule
\multirow{6}{*}{GPT-4o-mini}
& Extraversion & 0.255 & 0.177 & 0.078 & \cellcolor{red!40} -0.070 \\
& Agreeableness & 0.254 & 0.140 & 0.114 & \cellcolor{red!55} -0.106 \\
& Conscientiousness & 0.120 & 0.108 & 0.011 & \cellcolor{red!20} -0.004 \\
& Neuroticism & 0.162 & 0.120 & 0.041 & \cellcolor{red!30} -0.034 \\
& Openness & 0.222 & 0.136 & 0.087 & \cellcolor{red!45} -0.079 \\
& -- & 0.137 & 0.129 & 0.008 & 0.000 \\
\midrule
\multirow{6}{*}{Llama3-70B}
& Extraversion & 0.709 & 0.918 & -0.209 & \cellcolor{green!35} 0.035 \\
& Agreeableness & 0.648 & 0.907 & -0.260 & \cellcolor{red!25} -0.017 \\
& Conscientiousness & 0.376 & 0.578 & -0.202 & \cellcolor{green!40} 0.041 \\
& Neuroticism & 0.535 & 0.762 & -0.227 & \cellcolor{green!35} 0.017 \\
& Openness & 0.649 & 0.892 & -0.243 & \cellcolor{green!10} 0.000 \\
& -- & 0.460 & 0.703 & -0.243 & 0.000 \\
\midrule
\multirow{6}{*}{Llama3-8B}
& Extraversion & 1.000 & 0.952 & 0.048 & -- \\
& Agreeableness & 1.000 & 0.977 & 0.023 & -- \\
& Conscientiousness & 0.641 & 0.269 & 0.371 & -- \\
& Neuroticism & 0.889 & 0.050 & 0.840 & -- \\
& Openness & 1.000 & 0.975 & 0.025 & -- \\
\bottomrule
\end{tabular}
\caption{\textbf{Anchoring bias and mitigation effects of personality traits across models and traits.}  \textcolor{green!50}{Green-shaded values} represent a bias reduction. \textcolor{red!60}{Red-shaded values} indicate an increase in bias.  }
\end{table*}

\begin{table*}[htbp]
\centering
\footnotesize
\setlength{\tabcolsep}{4pt}

\renewcommand{\arraystretch}{1.2} 
\label{tab:framing-bias-mitigation}
\begin{tabular}{l l c c c c}
\toprule
Model & Trait & Admit Frame & Reject Frame & Framing Effect & Mitigation Effect \\
\midrule
\multirow{6}{*}{GPT-4o} 
& Extraversion & 0.023 & 0.021 & 0.002 & \cellcolor{green!50} 0.062 \\
& Agreeableness & 0.035 & 0.039 & -0.004 & \cellcolor{green!50} 0.060 \\
& Conscientiousness & 0.012 & 0.014 & -0.002 & \cellcolor{green!50} 0.062 \\
& Neuroticism & 0.025 & 0.027 & -0.002 & \cellcolor{green!50} 0.062 \\
& Openness & 0.029 & 0.037 & -0.008 & \cellcolor{green!48} 0.056 \\
& -- & 0.438 & 0.501 & -0.064 & 0.000 \\
\midrule
\multirow{6}{*}{GPT-4o-mini}
& Extraversion & 0.056 & 0.051 & 0.005 & \cellcolor{green!35} 0.008 \\
& Agreeableness & 0.056 & 0.060 & -0.004 & \cellcolor{green!37} 0.009 \\
& Conscientiousness & 0.035 & 0.030 & 0.005 & \cellcolor{green!35} 0.008 \\
& Neuroticism & 0.068 & 0.060 & 0.008 & \cellcolor{green!28} 0.005 \\
& Openness & 0.049 & 0.057 & -0.008 & \cellcolor{green!28} 0.005 \\
& -- & 0.041 & 0.054 & -0.013 & 0.000 \\
\midrule
\multirow{6}{*}{Llama3-70B}
& Extraversion & 0.358 & 0.442 & -0.084 & \cellcolor{red!55} -0.032 \\
& Agreeableness & 0.261 & 0.411 & -0.150 & \cellcolor{red!70} -0.098 \\
& Conscientiousness & 0.217 & 0.297 & -0.080 & \cellcolor{red!50} -0.028 \\
& Neuroticism & 0.132 & 0.313 & -0.181 & \cellcolor{red!75} -0.129 \\
& Openness & 0.421 & 0.514 & -0.093 & \cellcolor{red!60} -0.041 \\
& -- & 0.245 & 0.297 & -0.052 & 0.000 \\
\midrule
\multirow{6}{*}{Llama3-8B}
& Extraversion & 0.108 & 0.000 & 0.108 & \cellcolor{red!50} -0.086 \\
& Agreeableness & 0.072 & 0.000 & 0.072 & \cellcolor{red!50} -0.050 \\
& Conscientiousness & 0.019 & 0.000 & 0.019 & \cellcolor{green!25} 0.003 \\
& Neuroticism & 0.000 & 0.000 & 0.000 & \cellcolor{green!30} 0.022 \\
& Openness & 0.092 & 0.000 & 0.092 & \cellcolor{red!60} -0.070 \\
& -- & 0.022 & 0.000 & 0.022 & 0.000 \\
\bottomrule
\end{tabular}
\caption{\textbf{Framing bias and mitigation effects of personality traits across models.} \textcolor{green!50}{Green-shaded values} represent a bias reduction. \textcolor{red!70}{Red-shaded values} indicate an increase in bias.}
\end{table*}

\begin{table*}[htbp]
\centering
\footnotesize
\setlength{\tabcolsep}{4pt}

\renewcommand{\arraystretch}{1.2} 
\label{tab:decoy-effect}
\begin{tabular}{l l c c c c c}
\toprule
Model & Trait & Two-Option A & Decoy A & Decoy C & Decoy Effect & Mitigation Effect \\
\midrule
\multirow{6}{*}{GPT-4o} 
& Extraversion & 0.296 & 0.348 & 0.030 & 0.052 & \cellcolor{red!30} -0.016 \\
& Agreeableness & 0.033 & 0.185 & 0.023 & 0.152 & \cellcolor{red!55} -0.116 \\
& Conscientiousness & 0.153 & 0.381 & 0.027 & 0.228 & \cellcolor{red!60} -0.193 \\
& Neuroticism & 0.105 & 0.175 & 0.037 & 0.070 & \cellcolor{red!35} -0.035 \\
& Openness & 0.227 & 0.288 & 0.047 & 0.061 & \cellcolor{red!30} -0.026 \\
& -- & 0.335 & 0.371 & 0.062 & 0.036 & 0.000 \\
\midrule
\multirow{6}{*}{GPT-4o-mini}
& Extraversion & 0.286 & 0.115 & 0.000 & -0.172 & \cellcolor{green!60} 0.213 \\
& Agreeableness & 0.136 & 0.070 & 0.000 & -0.066 & \cellcolor{green!70} 0.318 \\
& Conscientiousness & 0.339 & 0.139 & 0.000 & -0.201 & \cellcolor{green!55} 0.184 \\
& Neuroticism & 0.225 & 0.118 & 0.000 & -0.107 & \cellcolor{green!65} 0.278 \\
& Openness & 0.255 & 0.122 & 0.000 & -0.132 & \cellcolor{green!63} 0.252 \\
& -- & 0.589 & 0.204 & 0.000 & -0.385 & 0.000 \\
\midrule
\multirow{6}{*}{Llama3-70B}
& Extraversion & 0.070 & 0.202 & 0.000 & 0.132 & \cellcolor{red!20} -0.004 \\
& Agreeableness & 0.016 & 0.078 & 0.088 & 0.061 & \cellcolor{green!40} 0.067 \\
& Conscientiousness & 0.161 & 0.236 & 0.000 & 0.075 & \cellcolor{green!35} 0.053 \\
& Neuroticism & 0.320 & 0.173 & 0.000 & -0.147 & \cellcolor{red!25} -0.019 \\
& Openness & 0.000 & 0.006 & 0.091 & 0.006 & \cellcolor{green!50} 0.122 \\
& -- & 0.300 & 0.428 & 0.000 & 0.128 & 0.000 \\
\midrule
\multirow{6}{*}{Llama3-8B}
& Extraversion & 0.000 & 0.249 & 0.020 & 0.249 & \cellcolor{red!55} -0.146 \\
& Agreeableness & 0.000 & 0.071 & 0.084 & 0.071 & \cellcolor{green!30} 0.032 \\
& Conscientiousness & 0.362 & 0.408 & 0.153 & 0.046 & \cellcolor{green!35} 0.057 \\
& Neuroticism & 0.000 & 0.021 & 0.054 & 0.021 & \cellcolor{green!45} 0.081 \\
& Openness & 0.000 & 0.046 & 0.099 & 0.046 & \cellcolor{green!35} 0.057 \\
& -- & 0.487 & 0.590 & 0.011 & 0.103 & 0.000 \\
\bottomrule
\end{tabular}
\caption{\textbf{Decoy effect and mitigation effects of personality traits across models.} \textcolor{green!50}{Green-shaded values} represent a bias reduction. \textcolor{red!70}{Red-shaded values} indicate an increase in bias.}
\end{table*}

\begin{table*}[htbp]
\centering
\footnotesize
\setlength{\tabcolsep}{4pt}

\renewcommand{\arraystretch}{1.2} 
\label{tab:risk-aversion}
\begin{tabular}{l l c c c c}
\toprule
Model & Trait & Neutral Rate & Averse Rate & Risk Effect & Mitigation Effect \\
\midrule
\multirow{6}{*}{GPT-4o} 
& Extraversion & 0.594 & 0.931 & 0.337 & \cellcolor{red!65} -0.293 \\
& Agreeableness & 0.000 & 0.323 & 0.323 & \cellcolor{red!63} -0.279 \\
& Conscientiousness & 0.000 & 0.166 & 0.166 & \cellcolor{red!50} -0.121 \\
& Neuroticism & 0.000 & 0.201 & 0.201 & \cellcolor{red!55} -0.157 \\
& Openness & 0.601 & 0.939 & 0.338 & \cellcolor{red!65} -0.294 \\
& -- & 0.015 & 0.059 & 0.044 & 0.000 \\
\midrule
\multirow{6}{*}{GPT-4o-mini}
& Extraversion & 0.609 & 0.999 & 0.390 & \cellcolor{green!55} 0.154 \\
& Agreeableness & 0.000 & 0.436 & 0.436 & \cellcolor{green!50} 0.108 \\
& Conscientiousness & 0.000 & 0.353 & 0.353 & \cellcolor{green!58} 0.192 \\
& Neuroticism & 0.000 & 0.382 & 0.382 & \cellcolor{green!55} 0.162 \\
& Openness & 0.603 & 1.000 & 0.397 & \cellcolor{green!53} 0.147 \\
& -- & 0.185 & 0.730 & 0.544 & 0.000 \\
\midrule
\multirow{6}{*}{Llama3-70B}
& Extraversion & 0.840 & 1.000 & 0.160 & \cellcolor{green!63} 0.268 \\
& Agreeableness & 0.000 & 0.210 & 0.210 & \cellcolor{green!60} 0.218 \\
& Conscientiousness & 0.000 & 0.000 & 0.000 & \cellcolor{green!75} 0.428 \\
& Neuroticism & 0.000 & 0.407 & 0.407 & \cellcolor{green!25} 0.021 \\
& Openness & 1.000 & 1.000 & 0.000 & \cellcolor{green!75} 0.428 \\
& -- & 0.097 & 0.525 & 0.428 & 0.000 \\
\midrule
\multirow{6}{*}{Llama3-8B}
& Extraversion & 0.172 & 0.581 & 0.409 & \cellcolor{red!70} -0.335 \\
& Agreeableness & 0.000 & 0.000 & 0.000 & \cellcolor{green!40} 0.074 \\
& Conscientiousness & 0.000 & 0.000 & 0.000 & \cellcolor{green!40} 0.074 \\
& Neuroticism & 0.000 & 0.172 & 0.172 & \cellcolor{red!45} -0.097 \\
& Openness & 0.551 & 0.895 & 0.344 & \cellcolor{red!63} -0.270 \\
& -- & 0.000 & 0.074 & 0.074 & 0.000 \\
\bottomrule
\end{tabular}
\caption{\textbf{Risk aversion and mitigation effects of personality traits across models.} \textcolor{green!50}{Green-shaded values} represent a bias reduction. \textcolor{red!70}{Red-shaded values} indicate an increase in bias.}
\end{table*}

\begin{table*}[htbp]
\centering
\footnotesize
\setlength{\tabcolsep}{4pt}

\renewcommand{\arraystretch}{1.2} 
\label{tab:sunk-cost-bias}
\begin{tabular}{l l c c c c}
\toprule
Model & Trait & Baseline Rate & Sunk Cost Rate & Sunk Cost Effect & Mitigation Effect \\
\midrule
\multirow{6}{*}{GPT-4o} 
& Extraversion & 0.000 & 0.008 & 0.008 & \cellcolor{red!35} -0.008 \\
& Agreeableness & 0.000 & 0.019 & 0.019 & \cellcolor{red!50} -0.019 \\
& Conscientiousness & 0.000 & 0.000 & 0.000 & 0.000 \\
& Neuroticism & 0.000 & 0.000 & 0.000 & 0.000 \\
& Openness & 0.000 & 0.000 & 0.000 & 0.000 \\
& -- & 0.000 & 0.000 & 0.000 & 0.000 \\
\midrule
\multirow{6}{*}{Llama3-70B}
& Extraversion & 0.000 & 0.000 & 0.000 & 0.000 \\
& Agreeableness & 0.000 & 0.000 & 0.000 & 0.000 \\
& Conscientiousness & 0.000 & 0.000 & 0.000 & 0.000 \\
& Neuroticism & 0.000 & 0.000 & 0.000 & 0.000 \\
& Openness & 0.000 & 0.000 & 0.000 & 0.000 \\
& -- & 0.000 & 0.000 & 0.000 & 0.000 \\
\midrule
\multirow{6}{*}{Llama3-8B}
& Extraversion & 0.000 & 0.000 & 0.000 & 0.000 \\
& Agreeableness & 0.000 & 0.000 & 0.000 & 0.000 \\
& Conscientiousness & 0.000 & 0.000 & 0.000 & 0.000 \\
& Neuroticism & 0.000 & 0.000 & 0.000 & 0.000 \\
& Openness & 0.000 & 0.000 & 0.000 & 0.000 \\
& -- & 0.000 & 0.000 & 0.000 & 0.000 \\
\bottomrule
\end{tabular}
\caption{\textbf{Sunk cost bias and mitigation effects of personality traits across models.} \textcolor{green!50}{Green-shaded values} represent a bias reduction. \textcolor{red!70}{Red-shaded values} indicate an increase in bias.}
\end{table*}

\begin{table*}[htbp]
\centering
\footnotesize
\setlength{\tabcolsep}{4pt}

\renewcommand{\arraystretch}{1.2} 
\label{tab:status-quo-bias}
\begin{tabular}{l l c c c c}
\toprule
Model & Trait & Status Quo Rate & Alternative Rate & Status Quo Effect & Mitigation Effect \\
\midrule
\multirow{6}{*}{GPT-4o} 
& Extraversion & 0.260 & 0.247 & 0.013 & \cellcolor{green!55} 0.120 \\
& Agreeableness & 0.330 & 0.223 & 0.107 & \cellcolor{green!30} 0.026 \\
& Conscientiousness & 0.392 & 0.203 & 0.189 & \cellcolor{red!35} -0.056 \\
& Neuroticism & 0.266 & 0.245 & 0.021 & \cellcolor{green!50} 0.112 \\
& Openness & 0.169 & 0.277 & -0.108 & \cellcolor{green!30} 0.025 \\
& -- & 0.350 & 0.217 & 0.134 & 0.000 \\
\midrule
\multirow{6}{*}{GPT-4o-mini}
& Extraversion & 0.163 & 0.279 & -0.116 & \cellcolor{red!25} -0.016 \\
& Agreeableness & 0.307 & 0.231 & 0.075 & \cellcolor{green!30} 0.025 \\
& Conscientiousness & 0.290 & 0.237 & 0.053 & \cellcolor{green!35} 0.048 \\
& Neuroticism & 0.331 & 0.223 & 0.108 & \cellcolor{red!20} -0.008 \\
& Openness & 0.144 & 0.285 & -0.142 & \cellcolor{red!35} -0.041 \\
& -- & 0.175 & 0.275 & -0.101 & 0.000 \\
\midrule
\multirow{6}{*}{Llama3-70B}
& Extraversion & 0.016 & 0.328 & -0.312 & \cellcolor{green!20} 0.008 \\
& Agreeableness & 0.080 & 0.307 & -0.226 & \cellcolor{green!45} 0.094 \\
& Conscientiousness & 0.058 & 0.314 & -0.257 & \cellcolor{green!40} 0.063 \\
& Neuroticism & 0.026 & 0.325 & -0.299 & \cellcolor{green!25} 0.021 \\
& Openness & 0.036 & 0.321 & -0.286 & \cellcolor{green!30} 0.034 \\
& -- & 0.010 & 0.330 & -0.320 & 0.000 \\
\midrule
\multirow{6}{*}{Llama3-8B}
& Extraversion & 0.214 & 0.262 & -0.048 & \cellcolor{red!35} -0.044 \\
& Agreeableness & 0.052 & 0.316 & -0.265 & \cellcolor{red!65} -0.261 \\
& Conscientiousness & 0.460 & 0.180 & 0.280 & \cellcolor{red!65} -0.277 \\
& Neuroticism & 0.124 & 0.292 & -0.168 & \cellcolor{red!60} -0.164 \\
& Openness & 0.090 & 0.268 & -0.178 & \cellcolor{red!60} -0.174 \\
& -- & 0.195 & 0.199 & -0.004 & 0.000 \\
\bottomrule
\end{tabular}
\caption{\textbf{Status quo bias and mitigation effects of personality traits across models.} \textcolor{green!50}{Green-shaded values} represent a bias reduction. \textcolor{red!70}{Red-shaded values} indicate an increase in bias.}
\end{table*}

\begin{table*}[htbp]
\centering
\footnotesize
\setlength{\tabcolsep}{4pt}

\renewcommand{\arraystretch}{1.2} 
\label{tab:endowment-effect}
\begin{tabular}{l l c c c c c c}
\toprule
Model & Trait & WTA & WTP & Control & WTA/WTP & Relative Effect (\%) & Relative Mitigation (\%) \\
\midrule
\multirow{6}{*}{GPT-4o} 
& Extraversion & 49923 & 14726 & 32116 & 3.39 & 109.59 & \cellcolor{red!85} -2467.85 \\
& Agreeableness & 0 & 2274 & 22562 & 0.00 & -10.08 & \cellcolor{red!55} -136.20 \\
& Conscientiousness & 28706 & 11916 & 29730 & 2.41 & 56.47 & \cellcolor{red!75} -1223.22 \\
& Neuroticism & 37302 & 10225 & 34276 & 3.65 & 79.00 & \cellcolor{red!80} -1750.92 \\
& Openness & 34818 & 14919 & 31177 & 2.33 & 63.83 & \cellcolor{red!77} -1395.51 \\
& -- & 28650 & 27042 & 37672 & 1.06 & 4.27 & 0.00 \\
\midrule
\multirow{6}{*}{GPT-4o-mini}
& Extraversion & 10357 & 7513 & 11534 & 1.38 & 24.67 & \cellcolor{red!25} -6.93 \\
& Agreeableness & 3140 & 2271 & 9084 & 1.38 & 9.57 & \cellcolor{green!40} 58.52 \\
& Conscientiousness & 8732 & 5319 & 11772 & 1.64 & 28.99 & \cellcolor{red!30} -25.70 \\
& Neuroticism & 8992 & 2827 & 9762 & 3.18 & 63.15 & \cellcolor{red!60} -173.79 \\
& Openness & 11423 & 7418 & 11935 & 1.54 & 33.56 & \cellcolor{red!35} -45.49 \\
& -- & 10439 & 7261 & 13777 & 1.44 & 23.07 & 0.00 \\
\midrule
\multirow{6}{*}{Llama3-70B}
& Extraversion & 13798 & 5377 & 11948 & 2.57 & 70.48 & \cellcolor{green!30} 22.74 \\
& Agreeableness & 3659 & 2204 & 4125 & 1.66 & 35.29 & \cellcolor{green!45} 61.32 \\
& Conscientiousness & 12420 & 4116 & 13518 & 3.02 & 61.43 & \cellcolor{green!35} 32.67 \\
& Neuroticism & 12317 & 835 & 9649 & 14.76 & 119.01 & \cellcolor{red!32} -30.45 \\
& Openness & 14442 & 5105 & 11804 & 2.83 & 79.10 & \cellcolor{green!25} 13.29 \\
& -- & 35475 & 10109 & 27806 & 3.51 & 91.23 & 0.00 \\
\midrule
\multirow{6}{*}{Llama3-8B}
& Extraversion & 16470 & 4265 & 13431 & 3.86 & 90.86 & \cellcolor{red!55} -128.47 \\
& Agreeableness & 3243 & 2281 & 3240 & 1.42 & 29.69 & \cellcolor{green!30} 25.34 \\
& Conscientiousness & 11206 & 4585 & 9688 & 2.44 & 68.35 & \cellcolor{red!45} -71.85 \\
& Neuroticism & 12774 & 2580 & 8496 & 4.95 & 119.99 & \cellcolor{red!70} -201.71 \\
& Openness & 19383 & 4610 & 11879 & 4.20 & 124.37 & \cellcolor{red!75} -212.72 \\
& -- & 19591 & 4767 & 37276 & 4.11 & 39.77 & 0.00 \\
\bottomrule
\end{tabular}
\caption{\textbf{Endowment effect and mitigation effects of personality traits across models.} \textcolor{green!50}{Green-shaded values} represent a bias reduction. \textcolor{red!70}{Red-shaded values} indicate an increase in bias. WTA = Willingness to Accept, WTP = Willingness to Pay.}
\end{table*}

\begin{table*}[htbp]
\centering
\footnotesize
\setlength{\tabcolsep}{4pt}

\renewcommand{\arraystretch}{1.2} 
\label{tab:group-attribution-bias}
\begin{tabular}{l l c c c c}
\toprule
Model & Trait & Male Rate & Female Rate & Group Attribution Bias & Mitigation Effect \\
\midrule
\multirow{6}{*}{GPT-4o} 
& Extraversion & 0.272 & 0.287 & -0.015 & \cellcolor{red!45} -0.013 \\
& Agreeableness & 0.275 & 0.285 & -0.010 & \cellcolor{red!35} -0.008 \\
& Conscientiousness & 0.234 & 0.239 & -0.005 & \cellcolor{red!25} -0.003 \\
& Neuroticism & 0.242 & 0.245 & -0.003 & \cellcolor{red!20} -0.001 \\
& Openness & 0.258 & 0.263 & -0.005 & \cellcolor{red!25} -0.003 \\
& -- & 0.247 & 0.249 & -0.002 & 0.000 \\
\midrule
\multirow{6}{*}{GPT-4o-mini}
& Extraversion & 0.279 & 0.301 & -0.022 & \cellcolor{red!30} -0.004 \\
& Agreeableness & 0.258 & 0.283 & -0.025 & \cellcolor{red!35} -0.007 \\
& Conscientiousness & 0.252 & 0.269 & -0.017 & \cellcolor{green!20} 0.001 \\
& Neuroticism & 0.270 & 0.293 & -0.023 & \cellcolor{red!30} -0.005 \\
& Openness & 0.267 & 0.291 & -0.024 & \cellcolor{red!35} -0.006 \\
& -- & 0.266 & 0.284 & -0.018 & 0.000 \\
\midrule
\multirow{6}{*}{Llama3-70B}
& Extraversion & 0.488 & 0.478 & 0.010 & \cellcolor{green!40} 0.009 \\
& Agreeableness & 0.222 & 0.218 & 0.004 & \cellcolor{green!45} 0.015 \\
& Conscientiousness & 0.138 & 0.132 & 0.006 & \cellcolor{green!43} 0.013 \\
& Neuroticism & 0.182 & 0.167 & 0.015 & \cellcolor{green!30} 0.004 \\
& Openness & 0.437 & 0.417 & 0.020 & \cellcolor{red!20} -0.001 \\
& -- & 0.294 & 0.275 & 0.019 & 0.000 \\
\midrule
\multirow{6}{*}{Llama3-8B}
& Extraversion & 0.243 & 0.232 & 0.011 & \cellcolor{red!40} -0.011 \\
& Agreeableness & 0.032 & 0.036 & -0.004 & \cellcolor{red!30} -0.004 \\
& Conscientiousness & 0.012 & 0.012 & 0.000 & 0.000 \\
& Neuroticism & 0.003 & 0.003 & 0.000 & 0.000 \\
& Openness & 0.041 & 0.057 & -0.016 & \cellcolor{red!45} -0.016 \\
& -- & 0.060 & 0.060 & 0.000 & 0.000 \\
\bottomrule
\end{tabular}
\caption{\textbf{Group attribution bias and mitigation effects of personality traits across models.} \textcolor{green!50}{Green-shaded values} represent a bias reduction. \textcolor{red!70}{Red-shaded values} indicate an increase in bias.}
\end{table*}

\end{document}